\documentclass[runningheads]{llncs}
\usepackage[T1]{fontenc}

\usepackage{amsmath}
\DeclareMathOperator*{\argmax}{argmax}

\usepackage{graphicx}
\usepackage{amsfonts}
\usepackage{soul}
\usepackage[small]{caption}

\usepackage{algorithm}
\usepackage{algorithmic}

\usepackage{array} 

\usepackage{multirow}
\usepackage{threeparttable}
\usepackage{adjustbox}
\usepackage[misc]{ifsym}

\begin{document}
\title{Pre-training Contextual Location Embeddings in Personal Trajectories via Efficient Hierarchical Location Representations}
\titlerunning{Geo-Tokenizer}
%
\author{Chung Park\inst{1,2}\and
Taesan Kim\inst{1}\and
Junui Hong\inst{1,2}\and
Minsung Choi\inst{1} \\ \and 
Jaegul Choo\inst{2}\Letter
}
\authorrunning{Park et al.}

\institute{SK Telecom, Seoul, Republic of Korea \\
\email{\{skt.cpark, ktmountain, skt.juhong, ms.choi\}@sk.com}\\
\and
Kim Jaechul Graduate School of AI, KAIST, Daejeon, Republic of Korea \\ \email{\{cpark88kr, secondrun3, jchoo\}@kaist.ac.kr}}
\tocauthor{Chung~Park,Taesan~Kim,Junui~Hong,Minsung~Choi,Jaegul~Choo}
\toctitle{Pre-training Contextual Location Embeddings in Personal Trajectories via Efficient Hierarchical Location Representations}

\maketitle
\begin{abstract}
Pre-training the embedding of a location generated from human mobility data has become a popular method for location based services.
In practice, modeling the location embedding is too expensive, due to the large number of locations to be trained in situations with fine-grained resolution or extensive target regions.
Previous studies have handled less than ten thousand distinct locations, which is insufficient  in the real-world applications.
To tackle this problem, we propose a Geo-Tokenizer, designed to efficiently reduce the number of locations to be trained by representing a location as a combination of several grids at different scales.
In the Geo-Tokenizer, a grid at a larger scale shares the common set of grids at smaller scales, which is a key factor in reducing the size of the location vocabulary.
The sequences of locations preprocessed with the Geo-Tokenizer are utilized by a causal location embedding model to capture the temporal dependencies of locations.
This model dynamically calculates the embedding vector of a target location, which varies depending on its trajectory.
In addition, to efficiently pre-train the location embedding model, we propose the Hierarchical Auto-regressive Location Model objective to effectively train decomposed locations in the Geo-Tokenizer.
We conducted experiments on two real-world user trajectory datasets using our pre-trained location model. 
The experimental results show that our model significantly improves the performance of downstream tasks with fewer model parameters compared to existing location embedding methods.

\end{abstract}

\section{Introduction}
\;\;\;\;\;\;For modeling human mobility patterns using large-scale mobility data, pre-training location embeddings using a self-supervised objective has advantages, because it allows comprehensive information about locations to be incorporated~\cite{lin2020pre}.
The pre-trained location embedding models can also be shared by a wide range of downstream models, such as those used for next location prediction or transportation mode classification, to improve the prediction performance as well as enhance computation efficiency \cite{wan2021pre}.

Many previous studies have applied language-modeling-based approaches to spatial-temporal datasets \cite{zhou2018deepmove,lin2020pre}.
For example, in DeepMove \cite{zhou2018deepmove}, the latent representations of places are trained by applying the skip-gram of word2vec \cite{mikolov2013efficient} to user trajectories.
CTLE \cite{lin2020pre} is a self-attention based location embedding model that considers a target location's contexts.
However, these previous studies still have limitations, as follows: First, the approaches are not scalable to real-world applications, which require numerous locations to be trained.
With the fine-grained resolution or extensive target regions, the number of distinct locations, the so-called \textbf{location vocabulary}, increases.
This deteriorates the quality and efficiency of the pre-trained embedding model because of the heavy embedding layer to be trained.
However, previous studies including Geo-Teaser \cite{zhao2017geo}, TrajFormer \cite{liang2022trajformer}, and CTLE \cite{lin2020pre} train the location embedding model with less than ten thousand locations.
Second, locations in a trajectory are often dependent on previously visited locations, meaning that the likelihood of visiting a specific location might be influenced by the locations stayed before \cite{lin2020pre,zhou2019context}. 
These dependencies can be short-term (e.g., dependencies between consecutive locations) or long-term (e.g., dependencies spanning multiple locations), and they are crucial factors for modeling the context-aware location embedding model.
However, previous studies have had difficulty capturing this sequential dependence between locations in their models.

\begin{figure}
  \begin{minipage}[c]{0.5\textwidth}
    \includegraphics[width=\textwidth]{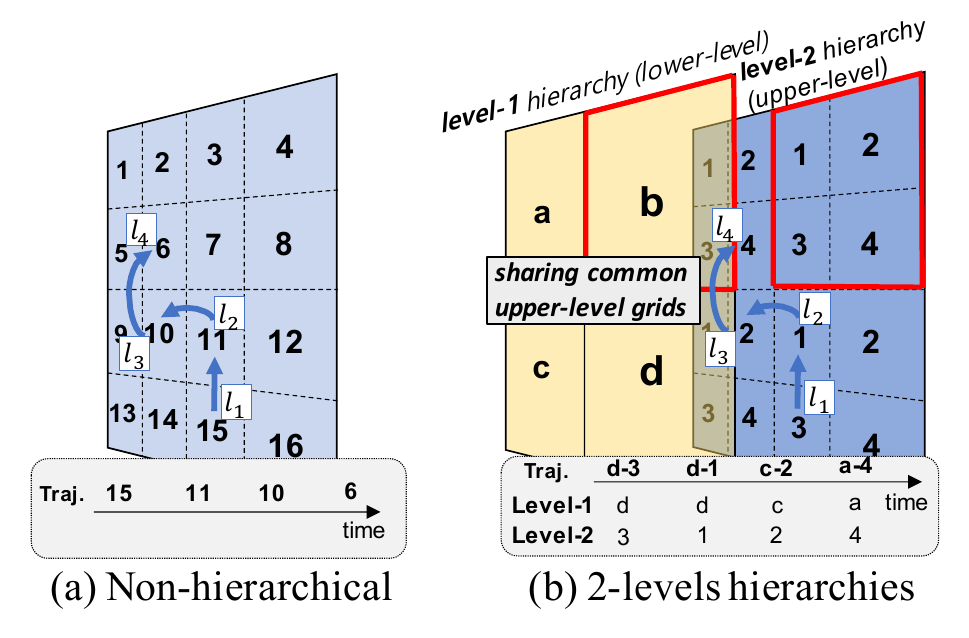}
  \end{minipage}\hfill
  \begin{minipage}[c]{0.5\textwidth}
    \caption{
    An illustration of spatial hierarchies at different scales ($H=2$). 
    (a) A trajectory with a non-hierarchical case is described.  
    (b) In our hierarchical case, each grid in the level-1 hierarchy shares the common grids set of $1,2,3,4$ in the level-2 hierarchy. For instance, $Grid\;6$ in the non-hierarchical case can be represented as ($Grid\;a$+$Grid\;4$) in the hierarchical case. 
    } \label{fig:method_hierarchy}
  \end{minipage}
\end{figure}

In order to tackle the discussed problems, we suggest a pre-trained location embedding model to efficiently handle numerous location vocabularies in various real-world applications. 
First, we devised the \textbf{geo-tokenizer embedding layer}, which represents a particular location as a combination of multiple grids at different scales to reduce the number of locations to be trained.
In this scheme, a specific location is represented as the combination of the $H$ tokens.
For example, as the location is composed of two hierarchies' grids in Figure \ref{fig:method_hierarchy}, its final representation is calculated by an element-wise sum of two hierarchies' grid embeddings.
\ul{Note that in our model, a grid in a lower (i.e., coarser-grained) hierarchy shares the common set of grids in upper (i.e., finer-grained) hierarchies, which is a key factor in reducing the location vocabulary size.}

Second, we designed a \textbf{causal location embedding model} consisting of the stack of the transformer decoder \cite{vaswani2017attention}. 
The transformer decoder inherently models temporal relationships due to its auto-regressive nature. 
This allows the model to capture the sequential patterns in the trajectory.
Therefore, we dynamically calculate the embedding of a target location considering its temporal order, which varies depending on its trajectory.

Lastly, to pre-train our location embedding model, we modified the Auto-regressive Language Model (ALM) objective introduced in the transformer \cite{vaswani2017attention}.
Since a grid in the lower hierarchy shares the those of the upper hierarchies in our model, specific two locations with far distance would have same lower-level (e.g., coarser-grained) embeddings despite that they may have different semantics or functionalities.
To solve this problem, we devised a \textbf{Hierarchical Auto-regressive Location Model (HALM)}.
This incorporated information from the lower-level hierarchies into the upper-level hierarchies when implementing ALM tasks to propagate the predicted output of lower-level hierarchies to the upper-level hierarchies. 
These components are incorporated in our model, as shown in Figure \ref{fig:pre-train_model}.
As a result, our location embedding model has relatively fewer parameters to learn and less computational cost than other competitive baselines.
In addition, it allows downstream task performance, such as next location prediction or transportation mode classification, to be improved with faster training and inference speed.

\section{Preliminaries}
\;\;\;\;\;\;A supplementary material (Appendix) with more details about the model, datasets and experiments is available at Github\footnote[1]{\url{https://github.com/cpark88/ECML-PKDD2023}}.

\noindent\textbf{Definition 1. Trajectory}: A trajectory is a sequence of locations where a person stays for a predefined time period \cite{shimizu2020learning,zhou2018trajectory}.
We set each location as a grid shape and $l_t$ as the $t$-th grid-shaped location.
Then, the sequence of visiting locations, denoted as a trajectory, can be defined as follows,

 \begin{equation}
 \label{equation:trajectory}
  s=\left\{l_0, l_1, \ldots, l_T\right\}
 \end{equation}
 where $T$ is the length of the trajectory $s$ and $l_0$ is the special token \texttt{SOS} which indicates the start of the trajectory. 
 We also denote $S$ as a set of trajectories.
We define the \textbf{location vocabulary} as the set of locations appearing in the train dataset, and denote it as $L$.
The size of $L$ is the vocabulary size of locations, denoted as $|L|$.

\noindent\textbf{Definition 2. Spatial Hierarchy}: 
Suppose that we set $H$-levels of \textbf{spatial hierarchies} $\{1,2,...,H\}$, where the level-$h$ hierarchy consists of grids with sizes of $r_h$ meters.
\ul{The uppermost hierarchy level $H$ has the smallest scale of $r_H$, thus the upper-level hierarchy has finer-grained grids than the lower-level.}

Each grid in the level-$(h$-$1)$ hierarchy is divided into a common grid set in the level-$h$ hierarchy, and the $t$-th location $l_t$ can be represented with a combination of $H$ grids at different scales.
From this spatial hierarchy, the location $l_t$ can be decomposed into the tuple of grids $(l^1_t, l^2_t, ..., l^H_t)$.
Therefore, trajectory $s$ consisting of decomposed locations from different hierarchies can be re-defined as follows,

 \begin{equation}
 \label{equation:trajectory_decomposed}
  s=\left\{(l^1_0, l^2_0, ..., l^H_0),(l^1_1, l^2_1, ..., l^H_1),\ldots,(l^1_T, l^2_T, ..., l^H_T)\right\}
 \end{equation}
 where $l^h_t$ is a level-$h$ grid at the $t$-th step and $l^h_0$ is the \texttt{SOS} token of the level-$h$ hierarchy.
 We define the set of all level-$h$ grids appearing in the train dataset as the location vocabulary of level-$h$ hierarchy, and denote it as $L^h$.
 The size of $L^h$ is denoted as $|L^h|$.
 Since $|L|\ge \sum_{h=1}^{H}|L^h|$, using the trajectories of decomposed locations is significantly efficient.
    
\begin{figure}[h]
    \begin{center}
    \includegraphics[width=0.87\linewidth]{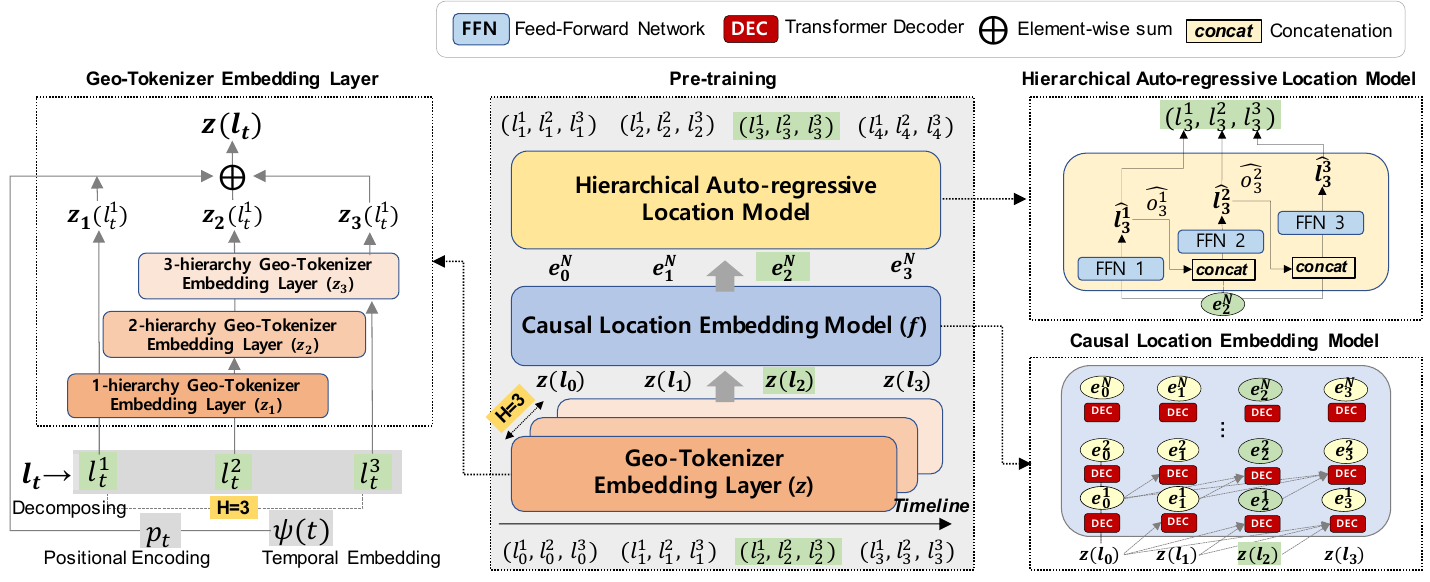}
    \end{center}
    \caption{We display the pre-training process of our model. The model with three level hierarchies(H) case is illustrated. 
    Note that $k(l_2)=f(z(l_0),z(l_1),z(l_2))=u(s_{<3})=e^{N}_{2}$ as described in Equation \ref{equation:pre-train_vector} and \ref{equation:final_mapping_func}.
    }
    \label{fig:pre-train_model}
    \end{figure}

\noindent\textbf{Problem Statement. Pre-training Location Embedding Model for Hierarchically Decomposed Locations}: 
Our goal is to pre-train a location embedding model $u$ to calculate a contextual embedding vector $k(l_{t})$ by predicting a next location $l_{t+1}=(l^1_{t+1}, l^2_{t+1}, ..., l^H_{t+1})$ given its context $s_{<t+1}=\{(l^1_0, l^2_0, ..., l^H_0),$
$(l^1_1, l^2_1, ..., l^H_1),\ \ldots,(l^1_{t}, l^2_{t}, ..., l^H_{t})\}$ with $H$ hierarchies.
We pre-train our model in a self-supervised manner as shown in Figure \ref{fig:pre-train_model}.

\section{Model}
\subsection{ Geo-tokenizer Embedding Layer}
\;\;\;\;\;\;We propose the  geo-tokenizer embedding layer, which allows the location embeddings to be trained  efficiently with a reduced number of location tokens.
By employing spatial hierarchies with different grid sizes, we can potentially capture varying levels of spatial patterns.
We first transform the input sequence into an embedding vector sequence.
As shown in Figure \ref{fig:pre-train_model}, we fetch an input latent representation $z(l_t)$ for $t$-th location $l_t$ from the embedding layer $z$. 
We call the embedding layer $z$ the geo-tokenizer embedding layer.
The embedding vector $z(l_t)$ can be described as follows,

 \begin{equation}
 \label{equation:geo_tokenizer}
  z(l_t)=(\sum_{h=1}^{H}z_{h}(l^h_{t}))+p_{t}+\psi(t)
 \end{equation}
where $z_h$ is a fully-connected embedding layer of the level-$h$ hierarchy, $l_{t}^h$ is a grid of the level-$h$ hierarchy at the $t$-th step, and $p_{t}$ is the $t$-th item of the positional encoding (PE) introduced in Transformer \cite{vaswani2017attention}.
The PE has an important role to capture the relative temporal position in the sequence.
In addition, inspired by the previous study \cite{li2017time}, we devised the temporal embedding $\psi(t)$.
For trajectories, the visiting records have temporal information which may significantly determine predicted locations.
$\psi(t)$ is calculated as follows,

 \begin{equation}
 \label{equation:temporal encoding}
  \psi(t)=\phi(log(r_t)W_{d}+b_d),
 \end{equation}
where $\phi$ is a nonlinear activation function (e.g., ReLU), and $r_t$ is an absolute timestamp at the $t$-th time step such as the real number in Unix Time.
$W_{d}$ is the trainable parameters for linearly transforming $log(r_t)$ and $b_d$ is the bias term.
The log transformation is conducted with $r_t$ to effectively cover the wide numerical range of temporal value \cite{li2017time}.
The dimension of $\psi(t)$ is equal to that of $z_{h}(l^h_t)$ and $p_t$.
This procedure generates an input sequence embedding $\{z(l_0),z(l_2)$ $,...,z(l_{t})\}$ for the causal location embedding model we will discuss in the next section.
Therefore, each embedding layer is represented by a matrix $z_{h}\in \mathbb{R}^{\vert L^{h}\vert \times W}$, where $\vert L^{h}\vert$ is the size of the vocabulary in the level-$h$ hierarchy, and $W$ is the embedding dimension. 

\subsection{Causal location embedding model}
\;\;\;\;\;\;The context of a target location can be obtained by the sequence of other locations before the target location in a trajectory.
From this perspective, we propose a causal location embedding model, which calculates a location’s latent representation by considering its contextual neighbors.
As shown in Figure \ref{fig:pre-train_model}, given a $(t+1)$-th target location $l_{t+1}$ and its context $s_{<t+1}$, we generate $t$-th location's final embedding vector $k(l_t)$ by using the casual location embedding model $f$ and the geo-tokenizer embedding layer $z$, denoted as follows:

 \begin{equation}
  \begin{split}
 \label{equation:pre-train_vector}
  &k(l_t)=f(z(l_0),z(l_1),...,z(l_{t})) \\
  &\;\;\;\;\;\;\;=u(l_0,l_1,...,l_{t}), \\
  &\;\;\;\;\;\;\;=u(s_{<t+1}),
 \end{split}
 \end{equation} 
where $u$ is our total location embedding model.
The embedding vector $k(l_t)$ is $t$-th item in output vectors' sequence of $u$.
Therefore, the embedding vector of $l_t$ is dynamically generated depending on the context $s_{<t+1}$.

The causal location embedding model $f$ consists of the stack of the transformer decoder \cite{vaswani2017attention}.
Due to the sequential nature of a trajectory, the model should take into account only the first $t$ items when predicting the $(t$+$1)$-th item.
This can consider the causal correlations of a target location and its contexts.
In addition, compared to the traditional sequential models such as LSTM \cite{hochreiter1997long}, it has the advantage of the long-term dependency and the parallelization with sequential datasets such as trajectories.
Also, unlike previous studies using the transformer encoder structure \cite{lin2020pre,park2021bertloc}, our model processes location information sequentially and can better handle both short-term and long-term dependencies in the trajectory (See the Appendix \ref{subsection:extended_ablation_study}).

Specifically, the input sequence embedding $\{z(l_0),z(l_2),...,z(l_{t})\}$ calculated in the geo-tokenizer embedding layer, is then fed into the causal location embedding model $f$, which is the stack of the transformer decoder.
A multi-head self-attention module with a causality mask and a feed-forward network are inherent in each transformer decoder \cite{vaswani2017attention}.
This process is described as: 
 \begin{equation}
 \begin{split}
 \label{equation:encoder}
 & \{\mathbf{e}^{(k)}_0,\mathbf{e}^{(k)}_1,...,\mathbf{e}^{(k)}_{t}\} \\
 & =\mathbf{Decoder}(\{\mathbf{e}^{(k-1)}_0,\mathbf{e}^{(k-1)}_1,...,\mathbf{e}^{(k-1)}_{t}\}), \\
 & \{\mathbf{e}^{(0)}_0,\mathbf{e}^{(0)}_1,...,\mathbf{e}^{(0)}_{t}\}=\{z(l_0),z(l_1),...,z(l_{t})\},
 \end{split}
 \end{equation}
where the \textbf{Decoder} represents the transformer decoder.
The output sequence of the $k$-th layer and the input sequence of the $(k$+$1)$-th layer are the same as $\{\mathbf{e}^{(k)}_0,\mathbf{e}^{(k)}_1,...,\mathbf{e}^{(k)}_{t}\}$.
We stack the $N$ transformer decoders in our causal location embedding module $f$.
The $t$-th item of the $N$-th transformer decoder is denoted as $\mathbf{e}^{N}_t$, which is the causal embedding vector of the location $l_t$.
In short, the final output vector of the location $l_t$ in the $N$ stack of the Decoder can be represented as:

 \begin{equation}
 \begin{split}
 \label{equation:final_mapping_func}
  k(l_t)=\mathbf{e}^{N}_t.
 \end{split}
 \end{equation}

\subsection{Pre-training Hierarchical Auto-regressive Location Model}
\;\;\;\;\;\;The relationship between target locations and their corresponding contexts should be considered in the location embedding model.
For this purpose, we propose the novel variant of the Auto-regressive Language Model (ALM) objective introduced in the transformer \cite{vaswani2017attention,radford2018improving,radford2019language}. 
In this paper, since we predict the next location in our pre-trained model, the ALM is rewritten as the Auto-regressive Location Model.
The ALM objective encourages the model to predict the next token with its context uni-directionally.
In this way, the correlation between the target token and its contexts can be captured in a self-supervised manner.
However, since a grid in the lower hierarchy shares the those of the upper hierarchies in our model, specific two locations with far distance would have same upper-level embeddings despite that they may have different semantics.
For this reason, we incorporated information from the lower-level hierarchies into the upper-level hierarchies when implementing ALM tasks to propagate the information of lower-level hierarchies to the upper-level hierarchies. 
In short, the predictions of upper-level hierarchies are contingent upon the predicted outcomes of lower-level hierarchies. 
This interdependence between hierarchical levels highlights the significance of integrating information across multiple scales to gain a comprehensive understanding of user trajectories.

As shown in Figure \ref{fig:pre-train_model}, we utilized a decomposed trajectory $s=\{l_0,l_1,\ldots,l_{t}\}=\{(l^1_0, l^2_0,\ldots, l^H_0),$
$(l^1_1, l^2_1,\ldots, l^H_1),\ldots,(l^1_{t}, l^2_{t},\ldots, l^H_{t})\}$ as the input of our location embedding model, and predicted the \textit{shifted} version of the input sequence $s$.
We train our model with multiple training objectives.
The ALM objectives of all hierarchies are trained simultaneously.
However, each ALM objective has a different task complexity.
Actually, since the grid size of the lower-level hierarchies is larger than that of the upper-level hierarchies, the trajectories of the lower-level hierarchies have monotonic patterns.
Therefore, the ALM objectives of lower hierarchies are much less demanding to train than the ALM objectives of upper hierarchies, which causes a learning imbalance between tasks.
In a multi-task architecture, the learning imbalance between tasks leads to causes the model to memorize a specific task instead of generalizing a pattern of data \cite{aksoy2020hierarchical}.
To solve this problem, in the ALM objectives, we sequentially incorporate the information from the lower hierarchy into the upper hierarchy.
We denote this multi-task objective as Hierarchical ALM (HALM).
The next location $l_{t+1}=(l^1_{t+1}, l^2_{t+1},\ldots, l^H_{t+1})$ to be predicted in the model consist of the $H$ tokens.
For this, we design $H$ fully-connected feed-forward networks to predict the $H$ next tokens using the causal location embedding model's output $\mathbf{e}_t^{(N)}$.
First, we predict $l^1_{t+1}$, the token of level-1 (i.e., the coarsest-grained) hierarchy in the $(t$+$1)$-step, as follows:
 \begin{equation}
 \begin{split}
 \label{equation:ffn}
    \widehat{l^1_{t+1}}=FFN^{1}_{HLM}(\mathbf{e}_{t}^{(N)}),
 \end{split}
 \end{equation}
where $FFN^{1}_{HLM}$ is the fully-connected feed-forward network of the level-1 hierarchy and $\widehat{l^1_{t+1}}$ is the prediction output for the next location token $l^1_{t+1}$.
In general, the prediction of the token of the level-$h$ hierarchy (i.e., $h$>1) in the $(t$+$1)$-step, is sequentially implemented as follows:
 \begin{equation}
 \begin{split}
 \label{equation:ffn_h}
    & \widehat{l^h_{t+1}}=FFN^{h}_{HLM}(\mathbf{e}_{t}^{(N)} \mathbin\Vert \widehat{o^{h-1}_{t+1}}), \\
    & \widehat{o^{0}_{t+1}}=\mathbf{0},
 \end{split}
 \end{equation}
where $\mathbin\Vert$ is the concatenation operation and $\widehat{o^{h-1}_{t+1}}$ is the one-hot encoding vector from prediction result $\widehat{l^{h-1}_{t+1}}$.
$FFN_{HLM}^{h}$ is composed of two fully-connected layers in this paper.
We construct the HALM objective to maximize the prediction accuracy of all of the hierarchies in the next location $l_{t+1}$.
The pre-training objective of the HALM can be described as:
 \begin{equation}
 \begin{split}
 \label{equation:objective}
O_{HALM}&= \argmax_{\theta} \sum_{h=1}^{H}\sum_{t=0}^{T}log(p(l^h_{t+1}|\widehat{l^h_{t+1}})),\\
 \end{split}
 \end{equation}
where $\theta$ denotes the set of all trainable parameters in our model, $T$ is the length of the trajectory, and $H$ is the uppermost level of the hierarchy (i.e., the finest-grained).

    \begin{figure*}[]
    \begin{center}
    \includegraphics[width=0.9\linewidth]{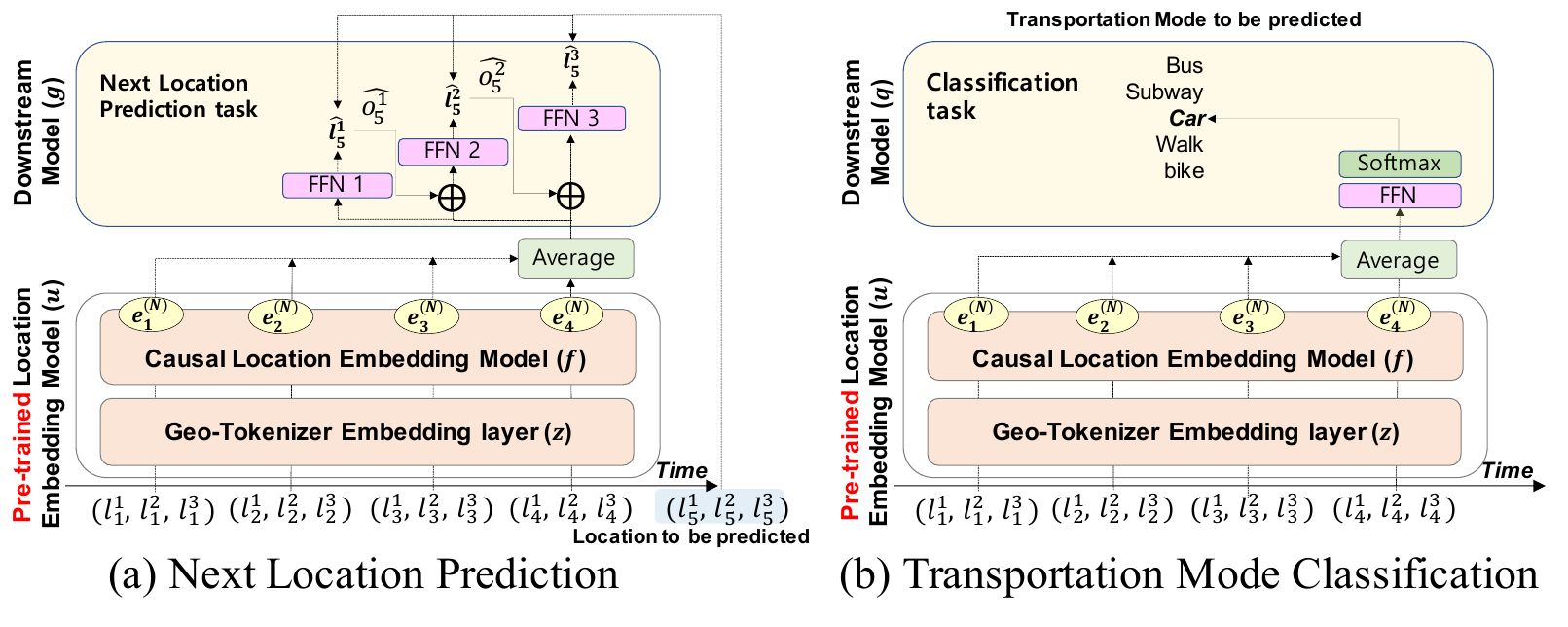}
    \end{center}
    \caption{
    Illustration of the downstream tasks.
    (a) Model architecture of the next location prediction task using an FFN layer stacked on the pre-trained location embedding model.
    (b) Model architecture of transportation mode classification task using one FFN layer stacked on the pre-trained location embedding model. 
    }
    \label{fig:finetune_model}
    \end{figure*}

\subsection{Fine-tuning Downstream tasks}  \label{ssec:Downstream}
\subsubsection{\textbf{Next Location Prediction task}} \label{sssec:NSP}
We implemented a next location prediction as a downstream task widely used in the location-based service in the real world \cite{zhao2020go,lin2020pre}.
The trajectory up to $T$ is used as an input in the downstream model, and the ground-truth is the three decomposed location records in $T+1$ (Figure \ref{fig:finetune_model}a).

\subsubsection{\textbf{Classification task}} \label{sssec:classification}
The model architecture for the transportation mode classification using a fully connected layer stacked on top of the pre-trained location embedding model is described in Figure \ref{fig:finetune_model}b.
A whole trajectory is used as an input in the downstream model, and the output is the transportation mode of the trajectory.
See Appendix \ref{ssec:Details of Fine-tuning} for details of above two downstream models.

\section{Experiments}
Our experiments are designed to answer the following research questions:

\noindent\textbf{(RQ1)}: How effective is our pre-trained location embedding model compared to the state-of-the-art models in the various downstream tasks?

\noindent\textbf{(RQ2)}: How do the different components affect the downstream tasks' performance?

\noindent\textbf{(RQ3)}: What is the effect of the level of hierarchies in our pre-trained model? 

\noindent\textbf{(RQ4)}: How effective is the pre-training of the location embedding model in the self-supervised manner on the downstream tasks? 


\begin{table*}[h]
\caption{Statistics of datasets}\label{tab:data_summary}
\begin{center}
\begin{adjustbox}{width=\columnwidth,center}
\begin{tabular}{cccccccccc}
\hline \hline
\multirow{2}{*}{Dataset} & \multirow{2}{*}{Data Type} & \multirow{2}{*}{\#Users} & \multirow{2}{*}{\begin{tabular}[c]{@{}c@{}}\#Original\\ Locations\\ (100m)\end{tabular}} & \multicolumn{4}{c}{\begin{tabular}[c]{@{}c@{}}\#Tokenized Locations (100m)\end{tabular}}                                                                                                                                                                     & \multirow{2}{*}{\#Traj} & \multirow{2}{*}{Time span} \\ \cline{5-8}
                         &                            &                          &                                                                                        & Total & \begin{tabular}[c]{@{}c@{}}\#Locations\\ level-1\\ (100km)\end{tabular} & \begin{tabular}[c]{@{}c@{}}\#Locations\\ level-2\\ (1km)\end{tabular} & \begin{tabular}[c]{@{}c@{}}\#Locations\\ level-3\\ (100m)\end{tabular} &                         &                            \\ \hline
\textbf{Mobile-T}               & Mobile Signal              & 0.4M                      & \textbf{79,812}                                                                                 & \textbf{6,740}  & 24                                                                                & 6,616                                                                            & 100                                                                              & 1.3M                     & 7/1,2021-7/31,2021         \\
\textbf{Geo-Life}                 & GPS                        & 182                      & \textbf{50,003}                                                                                  & \textbf{8,476}   & 183                                                                                & 8,193                                                                             & 100                                                                               & 17,621                  & 4/1,2007-8/31,2012          \\ \hline \hline
\end{tabular}
\end{adjustbox}
\end{center}
\end{table*}

\subsection{Datasets}
\noindent\textbf{Mobile-T}: 
This data is a set of user trajectories collected by the base stations of the major cellular network operator, denoted as Mobile-T.
As shown in Table \ref{tab:data_summary}, the size of location vocabularies at a 100m scale is 79812, which is too large to train for location embeddings.
However, using the Geo-tokenizer, the sizes of the location vocabularies in each hierarchy, 100km, 1km, and 100m scale, are 24, 6616, and 100, respectively.
\ul{This means that the total summation of the size of location vocabularies is 6740, which is less than 79812.}
Meanwhile, Mobile-T contains the land usage of the last location of a trajectory, associated with the purpose of the trajectory.
There are 15 unique land usages of a trajectory, such as Apartment House or Business Facilities.

\noindent \textbf{Geo-Life}\cite{zheng2010geolife}: We also used the public GPS trajectory dataset, Geo-Life, which was collected with 182 users over a period of five years in Microsoft Research Asia. 
In the Geo-Life dataset, the trajectories are described as sequences of locations represented
as GPS coordinates. 
Like the Mobile-T, the location record in this dataset was converted into a grid at a 100m scale.
\ul{In this dataset, the number of distinct decomposed locations with three hierarchies (8476) was less than the number of original locations (50003)}, as shown in Table \ref{tab:data_summary}. 
The Geo-life dataset contains five unique transportation modes of a trajectory.
See Appendix \ref{ssec:Details of Dataset} for details of two datasaets.

\subsection{Settings}
\;\;\;\;\;\;For both datasets, we assigned pre-train and fine-tune datasets of 80\% and 20\% of the total dataset.
Then, we assigned train, validation, and test datasets of 80\%, 10\%, and 10\% of the fine-tune datasets.
We trained fine-tuning (i.e., downstream) models with the train datasets and chose the optimal hyper-parameters with the validation datasets.
We set the hierarchy level $H$ as three, and the scales of the level-1, level-2, and level-3 hierarchies were 100km, 1km, and 100m, respectively.
We demonstrated the superiority of our pre-trained location model by comparing six location embedding models: (1) SERM \cite{yao2017serm}, (2) HIER \cite{shimizu2020learning}, (3) DeepMove \cite{zhou2018deepmove}, (4) TALE \cite{wan2019learning}, (5) CTLE \cite{lin2020pre}, and (6) TrajFormer \cite{liang2022trajformer}. 
Including our model, the dimension of the embedding layer and final location embedding vector was set to 256 in all models.
We described the model details in Appendix \ref{ssec:Details of baselines} and the pre-training setting in Appendix \ref{ssec:Training Details}.

\begin{table*}[]
\centering
\captionsetup{font=Large}
\begin{adjustbox}{width=\columnwidth,center}
\begin{threeparttable}[b]
\caption{Comparison of Next Location Prediction performance and efficiency with those of previous studies. The top two methods are highlighted in bold and underlined.}
\label{tab:result}

\begin{tabular}{cc|cc|cc|cccc}

\hline \hline 
\multicolumn{2}{c|}{\textbf{Downstream Model}}  

& \multicolumn{2}{c|}{\textbf{FFN}} & \multicolumn{2}{c|}{\textbf{LSTM}}   & \multirow{3}{*}{\textbf{\#Params}} & \multirow{3}{*}{\textbf{\begin{tabular}[c]{@{}c@{}}\#FLOPs\end{tabular}}} & 
\multirow{3}{*}{\textbf{\begin{tabular}[c]{@{}c@{}}TR-time\end{tabular}}} & \multirow{3}{*}{\textbf{\begin{tabular}[c]{@{}c@{}}Inf-time\end{tabular}}} \\ \cline{1-6}
\multicolumn{2}{c|}{\textbf{Metric}}                                          & \multicolumn{1}{c}{\multirow{2}{*}{\begin{tabular}[c]{@{}c@{}}Top-1\\ Acc(\%)\end{tabular}}} & \multicolumn{1}{c|}{\multirow{2}{*}{\begin{tabular}[c]{@{}c@{}}Top-5\\ Acc(\%)\end{tabular}}} & \multicolumn{1}{c}{\multirow{2}{*}{\begin{tabular}[c]{@{}c@{}}Top-1\\ Acc(\%)\end{tabular}}} & \multicolumn{1}{c|}{\multirow{2}{*}{\begin{tabular}[c]{@{}c@{}}Top-5\\ Acc(\%)\end{tabular}}} &      &         &           &         \\ \cline{1-2}
\multicolumn{1}{c|}{\textbf{Dataset}}     & \textbf{Pre-trained Model} & \multicolumn{1}{l}{}      & \multicolumn{1}{l|}{}     & \multicolumn{1}{l}{}      & \multicolumn{1}{l|}{}       &                &         &        &        \\ \hline
\multicolumn{1}{c|}{\multirow{7}{*}{\textbf{Mobile-T}}}  & SERM\cite{yao2017serm}      &  8.41$\pm$0.11  & 26.77$\pm$0.32      &   8.23$\pm$0.09   & 25.08$\pm$0.35     &   \underline{12.68M}      &  \underline{1.31B}      &  647.25    &   \ul{56.00}   \\
\multicolumn{1}{c|}{}     & HIER\cite{shimizu2020learning}      & 10.09$\pm$0.05      & 30.37$\pm$0.15    &  8.99$\pm$0.12    & 28.75$\pm$0.34      &    18.47M                                &   1.42B       &    837.39    &  71.47 \\
\multicolumn{1}{c|}{}     & DeepMove\cite{zhou2018deepmove}      & 9.05$\pm$0.15      & 30.81$\pm$0.19    &  9.38$\pm$0.30    & \underline{31.75$\pm$0.54}      &    49.92M                                &   2.62B       &    1338.47    &  116.55 \\
\multicolumn{1}{c|}{}       & TALE\cite{wan2019learning}        &  9.02$\pm$0.14  &  30.28$\pm$0.48    & \underline{9.43$\pm$0.14}      &  29.12$\pm$0.42    &   49.92M     &    7.85B      &     3493.91       &   285.57  \\
\multicolumn{1}{c|}{}         & CTLE\cite{lin2020pre}         &      \underline{10.71$\pm$0.24}   & \underline{32.39$\pm$1.09}         &       9.31$\pm$0.14    &  25.77$\pm$0.46        &     43.72M                       &   1.71B         &    \ul{642.61}            & 56.60  \\
\multicolumn{1}{c|}{}         & TrajFormer\cite{liang2022trajformer}         &   10.45$\pm$0.02      &   30.48$\pm$0.06      &   8.88$\pm$0.19        &    21.47$\pm$0.26      &         40.31M                  &    1.55B        &      693.37          &  56.20 \\
\multicolumn{1}{c|}{}         & \textbf{Ours}     &          \textbf{11.47$\pm$0.13}    & \textbf{38.41$\pm$0.11}          &            \textbf{11.20$\pm$0.05}     & \textbf{40.21$\pm$0.08}                    &       \textbf{6.90M}            &  \textbf{0.44B}     &    \textbf{400.63}                 & \textbf{43.37}    \\ \hline
\multicolumn{1}{c|}{\multirow{7}{*}{\textbf{Geo-Life}}}   & SERM\cite{yao2017serm}        &   18.35$\pm$0.11      &  29.02$\pm$0.18     &  18.58$\pm$0.31               &   36.46$\pm$0.44                  &   \underline{12.60M}                 &                 \underline{0.82B}                                                       &     \ul{187.12}              &    \ul{15.20}   \\
\multicolumn{1}{c|}{}     & HIER\cite{shimizu2020learning}      & 18.80$\pm$0.13      & 31.33$\pm$0.19    &  18.09$\pm$0.16    & \underline{38.34$\pm$0.13}      &    13.67M                                &   0.96B       &    207.43    &  16.37 \\
\multicolumn{1}{c|}{}            & DeepMove\cite{zhou2018deepmove}          &  17.46$\pm$0.11       &  35.72$\pm$0.12       &   18.68$\pm$0.19        &  38.03$\pm$0.22   &     25.60M           &     1.64B      &         345.66      &    30.38   \\
\multicolumn{1}{c|}{}             & TALE\cite{wan2019learning}              &  17.58$\pm$0.13     &    31.17$\pm$0.18        &             19.04$\pm$0.12                     &  38.15$\pm$0.14         &     25.60M                   &      4.92B   &               678.68  & 59.59   \\
\multicolumn{1}{c|}{}        & CTLE\cite{lin2020pre}              &          24.69$\pm$0.25      & 45.12$\pm$0.15       &        21.53$\pm$0.38      &  43.49$\pm$0.36                           &        30.39M      &   1.18B      &           192.10       &     16.48     \\
\multicolumn{1}{c|}{}         & TrajFormer\cite{liang2022trajformer}         &   \ul{27.71$\pm$0.97}      & \ul{50.51$\pm$0.93}         &   \ul{26.34$\pm$0.19}        &  \ul{51.53$\pm$0.30}        &           28.91M                &       1.06B      &       223.44         &  16.97 \\
\multicolumn{1}{c|}{}         & \textbf{Ours}     &        \textbf{28.58$\pm$0.21}      & \textbf{60.07$\pm$0.31}         &          \textbf{26.99$\pm$0.28}      & \textbf{53.68$\pm$0.61}                       &    \textbf{7.71M}           &     \textbf{0.48B}        &    \textbf{179.92}             &  \textbf{14.74} \\ \hline \hline
\end{tabular}
  \begin{tablenotes}
    \item[*] The number of parameters and FLOPs are derived from only location embedding models, except downstream task models (FFN and LSTM).
    M and B denote million and billion, respectively.
    We executed each baseline ten times and recorded the mean and standard deviation of each baseline.
    TR-time and Inf-time indicate the training time (seconds) per epoch and inference time (seconds) per epoch in the FFN case of the next location prediction model respectively.
    The training and inference speed were calculated by averaging those of FFN and LSTM with each pre-trained model, using one V100 GPU.
    \end{tablenotes}
\end{threeparttable}
\end{adjustbox}
\end{table*}

\subsection{Experimental Results (RQ1)}
\subsubsection{Next Location Prediction task}
The performance of the next location prediction task was assessed using the accuracy of the test dataset.
The rate at cutoff $k$, denoted as \textbf{Acc@\textit{k}}, counts the fraction of cases where the target location is among the top $k$.
We reported this metric as $k$=1 and $k$=5.
We also evaluated the efficiency of the pre-trained location embedding model by measuring the number of model parameters and operations (FLOPs).
A performance comparison for the next location prediction task is shown in Table \ref{tab:result}.
SERM \cite{yao2017serm} with the randomly initialized embedding layers did not perform well because this method has difficulty incorporating the context of a trajectory.
DeepMove \cite{zhou2018deepmove} and TALE \cite{wan2019learning} adopting Skip-gram and CBOW utilize the co-occurrence probabilities of target locations and their contexts, but the contexts they consider were restricted to specific window size.
More importantly, these methods train the heavy embedding layers due to the large size of the location vocabulary, which was over 50,000 in both datasets.

Unlike the above previous studies, CTLE \cite{lin2020pre} and TrajFormer \cite{liang2022trajformer} incorporate the multi-functionality of a location via a self-attention module to consider the contexts of trajectories.
As a result, they showed significantly better performance than the other baseline models.
However, they also had difficulty dealing with the large size of the location vocabulary and needed to train the heavy embedding layers.
\ul{Our model consistently outperformed other location embedding methods, even with fewer parameters and the number of FLOPs.} 
This can be attributed to the efficient processing of large amounts of location vocabulary using the Geo-tokenizer embedding layer and HALM objective. 
\ul{In addition, our model was faster than other baselines in the training and inference for both datasets} (Table \ref{tab:result}).

Concurrently, our experimental results demonstrate that, for the next location prediction tasks, adopting a feed-forward network in conjunction with a self-attention-based pre-trained model (e.g., CTLE\cite{lin2020pre}, TrajFormer\cite{liang2022trajformer}, Ours) proves to be more competitive than utilizing an LSTM-based approach. 
One potential reason is that the pre-trained model, which is based on a self-attention layer, has already learned to capture long-range dependencies in the input sequence. 
In this case, adding another layer of sequential processing with LSTM may not provide significant additional benefits. 

\subsubsection{Classification task}
The performance of the land usage and transportation mode classification task was assessed using the accuracy, macro-precision, and macro-recall of the test dataset.
A performance comparison for these tasks is shown in Table \ref{tab:result_classfier}.
\ul{With the fewest parameters, our pre-trained location embedding model showed the best performances among other location embedding models for both tasks, and was faster than other baselines in the training and inference.}
This indicates the superior quality of our pre-trained location embeddings. 

\begin{table*}[]
\centering
\begin{adjustbox}{width=\columnwidth,center}
\begin{threeparttable}[b]
\captionsetup{font=Large}
\caption{Comparison of Land Usage and Transportation Mode Classification task with those of previous studies. The top two methods are highlighted in bold and underlined.}
\label{tab:result_classfier}
\begin{tabular}{c|c|c|c|c|c|c|c|c}
\hline
\hline
\multirow{2}{*}{\textbf{Dataset}} & \multirow{2}{*}{\textbf{\begin{tabular}[c]{@{}c@{}}Downstream\\ Task\end{tabular}}}                                                     & \textbf{Metric}   & \multirow{2}{*}{\textbf{Accuracy(\%)}} & \multirow{2}{*}{\textbf{Precision(\%)}} & \multirow{2}{*}{\textbf{Recall(\%)}} & \multirow{2}{*}{\textbf{F-1(\%)}} & \multirow{2}{*}{\textbf{\begin{tabular}[c]{@{}c@{}}TR-time\end{tabular}}} & \multirow{2}{*}{\textbf{\begin{tabular}[c]{@{}c@{}}Inf-time\end{tabular}}} \\ \cline{3-3}
                                  &                                                                                               & \textbf{Pre-trained Model} &                                        &                                         &                                      &                                   &                                                                                       &                                                                                        \\ \hline
\multirow{7}{*}{\textbf{Mobile-T}}         & \multirow{7}{*}{\begin{tabular}[c]{@{}c@{}}Land Usage\\ Classification\end{tabular}}          & SERM\cite{yao2017serm} &         79.12$\pm$0.02                      &               73.83$\pm$0.03                 &           70.35$\pm$0.02                    &         70.65$\pm$0.02     &  \ul{58.23} &    \ul{5.42}        \\
                            &                                                     & HIER\cite{shimizu2020learning}          &        79.39$\pm$0.05     
                            &                 76.70$\pm$0.03               &          73.31$\pm$0.03                  &         74.42$\pm$0.02       & 121.28 &   6.94       \\
                            &                                                     & DeepMove\cite{zhou2018deepmove}          &        81.05$\pm$0.09      
                            &                 \ul{77.95$\pm$0.02}               &          \ul{73.43$\pm$0.03}                  &         75.38$\pm$0.02       & 129.09 &   10.99       \\                            
                            &                                                     & TALE\cite{wan2019learning}              &             83.02$\pm$0.06                   &        77.09$\pm$0.07                        &           73.33$\pm$0.10                   &         \ul{76.47$\pm$0.11}       & 310.95 &  26.33         \\
                            &                                                     & CTLE\cite{lin2020pre}               &      \ul{87.44$\pm$1.29}                         &     75.42$\pm$2.97                           &          73.31$\pm$2.37                   & 73.22$\pm$2.04           & 58.36 & 4.98             \\

                            & & TrajFormer\cite{liang2022trajformer}               &    73.65$\pm$0.83                           &      67.37$\pm$4.90                          &   56.24$\pm$1.36                          &  59.42$\pm$2.08          & 66.86 &    6.68          \\

                            &                                                     & \textbf{Ours}     &        \textbf{89.47$\pm$0.29}                      &        \textbf{82.27$\pm$1.65}                        &       \textbf{84.19$\pm$0.94}                      &      \textbf{82.80$\pm$0.95}         & \textbf{41.72} & \textbf{4.51}                                                                               \\ \hline
\multirow{7}{*}{\textbf{Geo-Life}}         & \multirow{7}{*}{\begin{tabular}[c]{@{}c@{}}Transportation\\ Mode \\ Classification\end{tabular}} & SERM\cite{yao2017serm} &          68.19$\pm$0.02                      &              69.13$\pm$0.03                  &          69.21$\pm$0.03                    &      69.19$\pm$0.03       & \ul{5.26} &  \ul{0.58}            \\
                            &                                                     & HIER\cite{shimizu2020learning}          &      64.45$\pm$0.17       &        60.56$\pm$0.12                 &             64.52$\pm$0.21                    &         64.44$\pm$0.14                      &          6.44 & 0.74            \\
                            &                                                     & DeepMove\cite{zhou2018deepmove}          &              69.81$\pm$0.09                 &             71.46$\pm$0.02                    &         69.96$\pm$0.07                      &         69.95$\pm$0.08       & 9.58 & 1.28            \\                            
                            &                                                     & TALE\cite{wan2019learning}              &               62.88$\pm$0.08                &          70.32$\pm$0.09                       &      65.86$\pm$0.11                       &          66.53$\pm$0.11    & 18.78 &  2.51           \\
                            &                                                     & CTLE\cite{lin2020pre}              &      68.15$\pm$1.10                         &      71.36$\pm$1.10                          &           73.21$\pm$0.84                  &        71.01$\pm$1.06        & 5.30 & 0.72         \\
                            & & TrajFormer\cite{liang2022trajformer}               &  \ul{73.68$\pm$1.88}                             &   \ul{77.37$\pm$1.33}                             &     \ul{76.49$\pm$1.95}                         &  \ul{76.10$\pm$1.68}         &5.88  &       0.59       \\
                            &                                                     & \textbf{Ours}     & \textbf{81.17$\pm$0.40}                     &                 \textbf{81.58$\pm$0.74}               &    \textbf{82.70$\pm$0.75}                          &     \textbf{81.81$\pm$0.32}                                   &  \textbf{4.34}                                                                                     &        \textbf{0.46}                                                                                \\ \hline\hline
\end{tabular}
  \begin{tablenotes}
    \item[*] The number of parameters and FLOPs in each pre-trained embedding model is equal to the case of the next location prediction task, as shown in Table \ref{tab:result}. We executed each baseline ten times and recorded the mean and standard deviation of each baseline.
    \end{tablenotes}
\end{threeparttable}
\end{adjustbox}
\end{table*}

\subsection{Ablation study} \label{subsection:ablation_study}
\subsubsection{Study on the components (RQ2)}
 We investigated the effectiveness of each component of our pre-trained location embedding model by designing three variants as follows:

\noindent\textbf{(1) Baseline}: 
This model utilizes the original transformer decoder using the ALM objective for pre-training without the Geo-tokenizer embedding layer.
This is a simple auto-regressive pre-trained model.

\noindent\textbf{(2) +Geo-tokenizer(GT)}: 
This model replaces the embedding layers in the baseline with the Geo-tokenizer embedding layer, which decomposes each location record into the three hierarchical components (100km, 1km, 100m). 
The pre-trained model's objective is the basic ALM proposed in the transformer \cite{vaswani2017attention}.
Therefore, the ALM objectives of the three hierarchies are independent.

\noindent\textbf{(3) +Geo-tokenizer(GT)+HALM}: 
This model uses the Geo-tokenizer fused on the baseline and employs the HALM objective.
This is our proposed model.

\begin{figure*}[h]
\begin{minipage}[t]{0.5\columnwidth}
    \centering
    \includegraphics[width=1\linewidth]{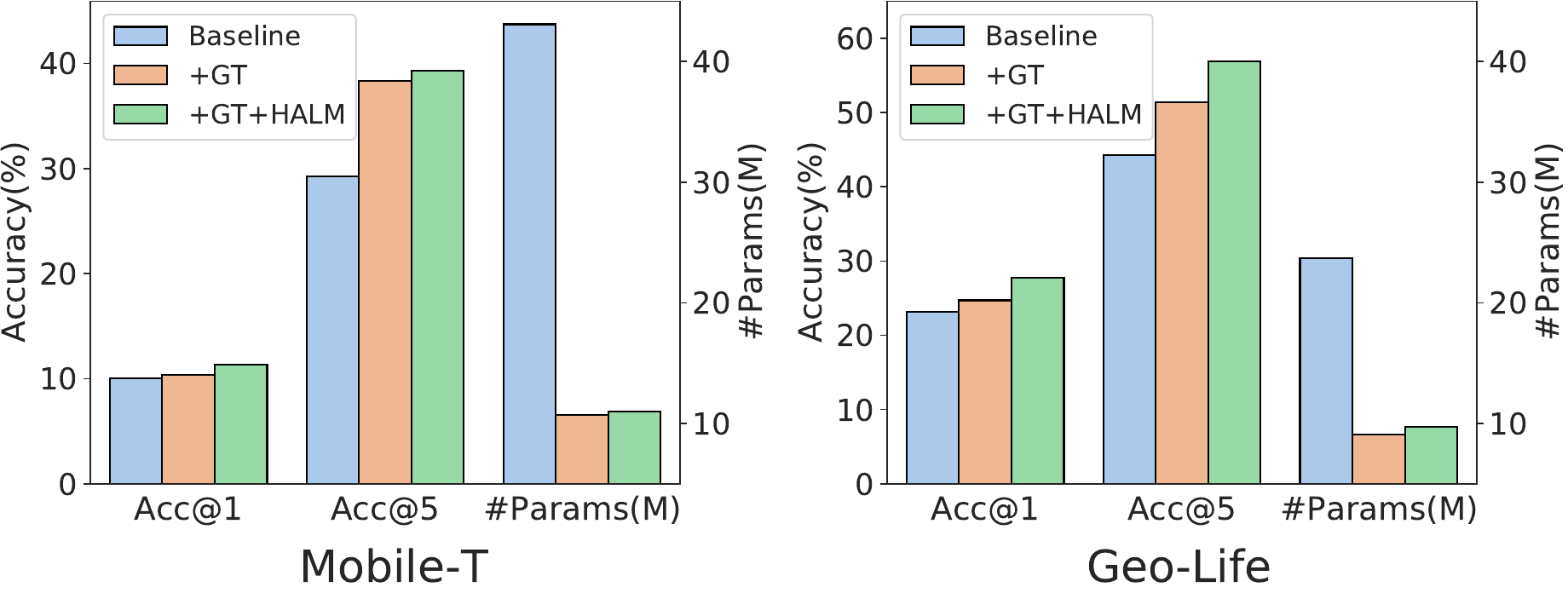}
    \caption{Comparison of next location prediction performance and efficiency for different combinations of components. 
        }
    \label{fig:ablation_2}
\end{minipage}
\hspace{0.1cm}
\begin{minipage}[t]{0.5\columnwidth} 
    \centering
    \includegraphics[width=1\linewidth]{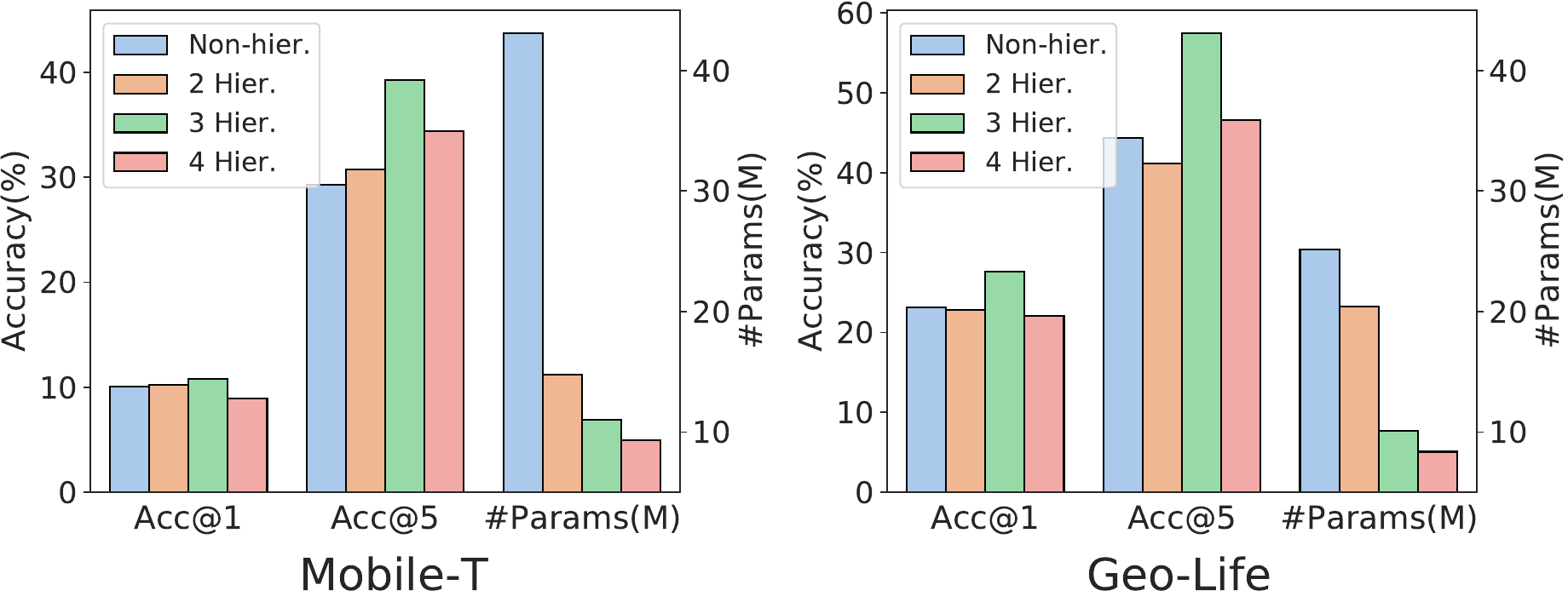}
    \caption{Next location prediction performance and efficiency comparison of different hierarchy levels. 
    }
    \label{fig:ablation}
\end{minipage}        
\end{figure*}

\begin{figure*}[h]
\begin{minipage}[t]{0.5\columnwidth}
    \centering
    \includegraphics[width=1\linewidth]{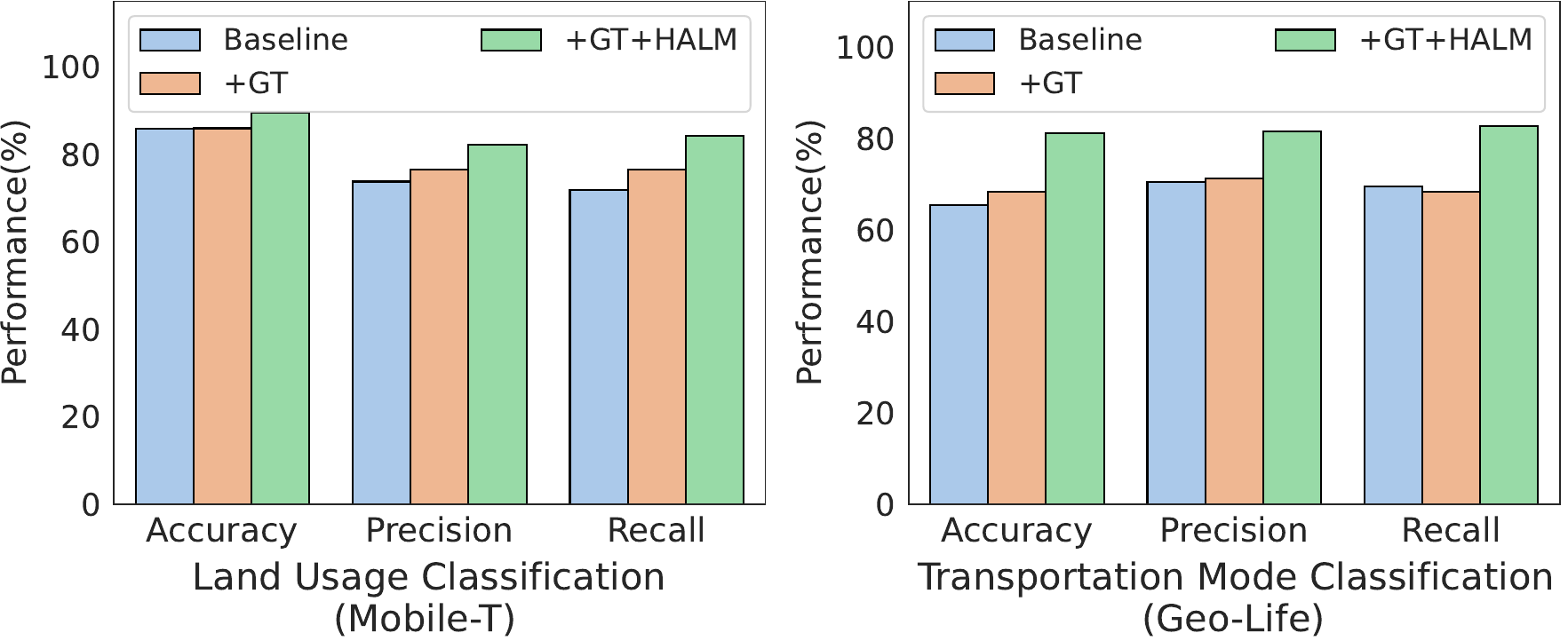}
    \caption{Comparison of classification performance and efficiency for different combinations of components.
        }
    \label{fig:ablation_classification_1}
\end{minipage}
\hspace{0.1cm}
\begin{minipage}[t]{0.5\columnwidth} 
    \centering
    \includegraphics[width=1\linewidth]{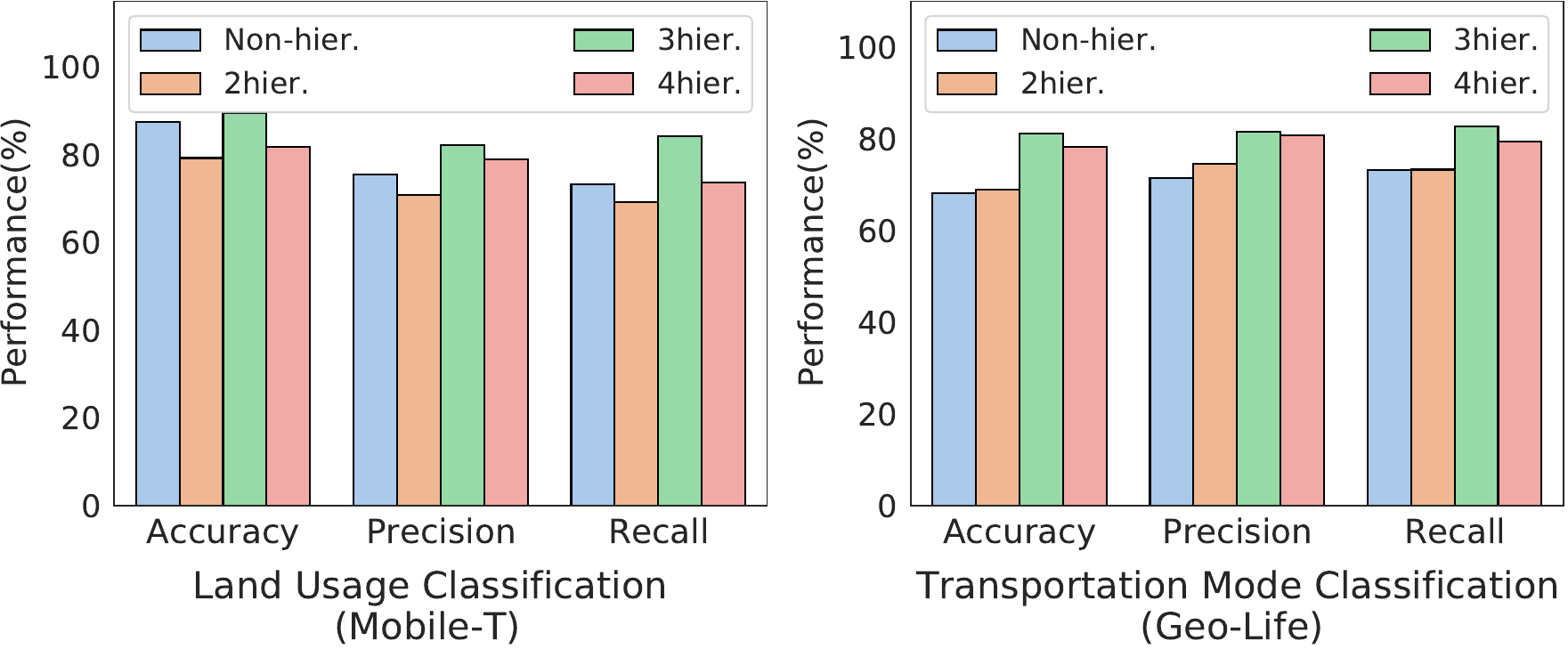}
    \caption{Comparison of classification performance and efficiency for different hierarchy levels.
    }
    \label{fig:ablation_classification_2}
\end{minipage}        
\end{figure*}

The comparison of these three variants was conducted with our pre-trained location embedding model on the next location prediction task, shown in Figure \ref{fig:ablation_2}.
The performance was calculated by averaging two downstream models (FFN and LSTM).
Compared to the baseline model, the model with the Geo-tokenizer embedding layer showed higher performance in both datasets.
In addition, the model combining the HALM objective with the Geo-tokenizer embedding layer outperformed other variants.
\ul{This means that the learning imbalance caused by location decomposition into multiple hierarchies by the Geo-tokenizer embedding layer was resolved through HALM.}
The comparison of these three variants was conducted with our pre-trained location embedding model on the classification tasks shown in Figure \ref{fig:ablation_classification_1}.
In the land usage and transportation mode classification tasks, both the Geo-tokenizer embedding layer and HALM can improve the prediction performance over the baseline.

\subsubsection{Study on the level of hierarchies (RQ3)}
We studied the effectiveness of the level of hierarchies by comparing three variants in terms of the degree of hierarchies: 
\noindent\textbf{(1) Four} (100km,10km, 1km, and 100m),
\noindent\textbf{(2) Three} (100km, 1km, and 100m), and
\noindent\textbf{(3) Two hierarchies case} (10km and 100m).

The performance was calculated by averaging those of FFN and LSTM.
As shown in Figure \ref{fig:ablation}, the three hierarchies case showed the best Acc@\textit{1} and Acc@\textit{5} with relatively few parameters for both datasets on the next location prediction task.
In addition, we determined that increasing the hierarchy level did not necessarily improve the next location prediction performance.
The larger the hierarchy level($H$), the smaller the location vocabulary size, resulting in a smaller model size. 
If the model size is too small, the performance deteriorates, so it can be seen that setting an appropriate $H$ is essential.
We also compared these three variants with the non-hierarchies model on the two classification tasks, as shown in Figure \ref{fig:ablation_classification_2}.
In the both classification tasks, the three hierarchies case showed the best performance with the fewest parameters.
\ul{It can be seen that the performance of the hierarchical case above a certain level is better than that of the non-hierarchical case with fewer model parameters.}

\subsubsection{Study on the pre-training (RQ4)}
\begin{figure*}[h]
\begin{minipage}[t]{0.5\columnwidth}
    \centering
    \includegraphics[width=1\linewidth]{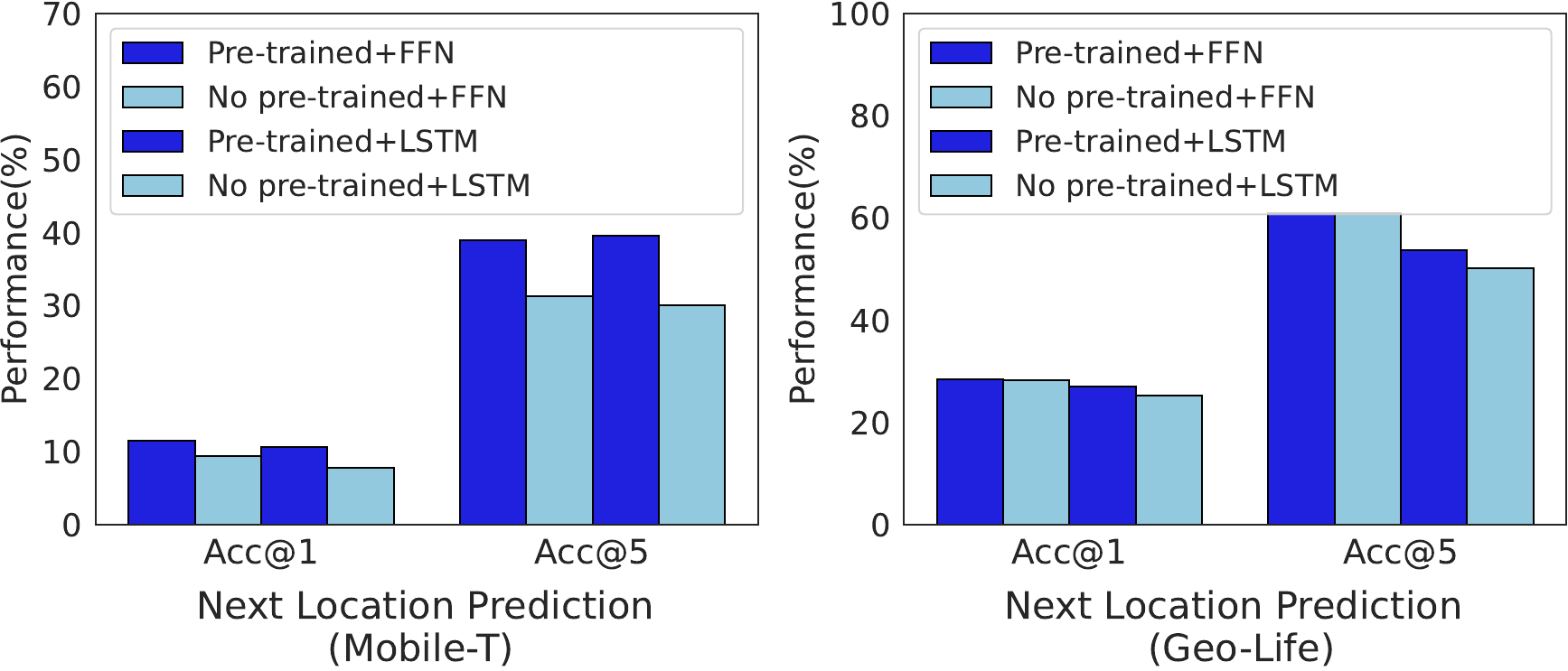}
    \caption{Effect on the pre-training for the next location prediction task.
        }
    \label{fig:pre-train_nlp}
\end{minipage}
\hspace{0.1cm}
\begin{minipage}[t]{0.5\columnwidth} 
    \centering
    \includegraphics[width=1\linewidth]{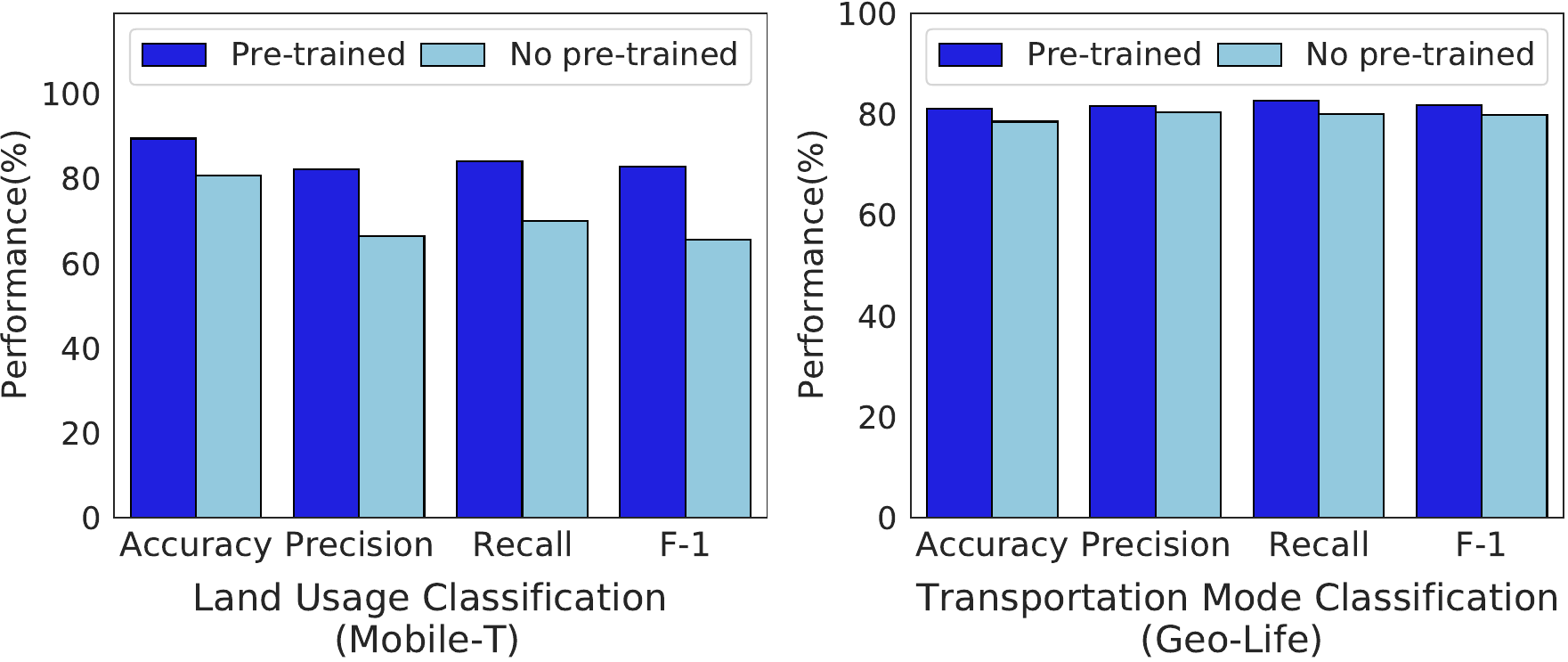}
    \caption{Effect on the pre-training for the classification task.
    }
    \label{fig:pre-train_classification}
\end{minipage}        
\end{figure*}

\ul{Pre-training significantly improved the performance of downstream tasks.}
We compared the performance of our model in two cases: with pre-training (w/PT) and without pre-training (wo/PT). 
As shown in Figure \ref{fig:pre-train_nlp}, the model with the pre-trained backbone showed higher performance in both datasets for the next location prediction task than the wo/PT.
In the Geo-Life dataset, the performance gap between the w/PT and wo/PT was relatively small compared to that of Mobile-T.
This is because the number of trajectories of Mobile-T is larger than Geo-Life's.
In other words, the larger the data, the greater the performance improvement of the downstream task due to pre-training.
In the classification task, the w/PT performed significantly better than the wo/PT in terms of accuracy, precision, and recall in both datasets, as shown in Figure \ref{fig:pre-train_classification}.

\subsection{Deployed Solution}
\;\;\;\;\;\;Our pre-trained model has been implemented in an inter-company marketing tool, designed to predict individuals likely to visit a particular area for location-based marketing purposes. The deployed solution effectively encompasses entire regions within the author's country by utilizing the next-location prediction model built upon our pre-trained model.
More details, including a screenshot of our graphical user interface (GUI) tool, can be found in the Appendix \ref{sec:Our Deployed Solution}.

\section{Related Work} \label{sec:related_work}
\;\;\;\;\;\;In recent years, pre-training an embedding model with self-supervised objectives has become a common practice in spatial-temporal data mining.
For example, DeepMove \cite{zhou2018deepmove} and TALE \cite{wan2019learning} implemented skip-gram and CBOW  \cite{mikolov2013efficient}, respectively, to model human mobility, and an N-gram model is adopted to learn latent representations of a location \cite{yao2018representing,shimizu2020learning}. 
SERM \cite{yao2017serm} jointly trained the embeddings of user, location, time, and keyword. 
These location embedding models generated a single latent representation for each location, which indicates they can not discriminate among variable functionalities of a location.
To address this problem, previous studies have employed a transformer encoder architecture \cite{vaswani2017attention} with Masked Language Model\cite{devlin-etal-2019-bert} to generate the dynamic embeddings derived along the dissimilar trajectories \cite{park2021bertloc,lin2020pre}.
Specifically, TrajFormer \cite{liang2022trajformer}, CTLE \cite{lin2020pre} and BERTLoc \cite{park2021bertloc} proposed a transformer encoder based location embedding model that dynamically assigns the embedding vector of a target location, varying with the location's trajectory.
 Nevertheless, previous studies are difficult to be applied in the real world, where the number of locations can be considerably large, or a fine-grained resolution is needed~\cite{shimizu2020learning}.
Previous studies have dealt with at most ten thousand locations to train their representations \cite{lin2020pre,shimizu2020learning,zhao2017geo}.
This problem can be addressed by reconstructing a location with several grids at different scales and making each grid at a large scale share the grids at a small scale.
HIER \cite{shimizu2020learning} decomposed a location at several spatial scales to consider the spatial hierarchy in the location embeddings. 
However, in their approaches, locations in each level of the hierarchy are independently trained, and therefore the number of locations to be embedded is still large. 
For this reason, we encourage grids in the lower-level hierarchies to share the grid set in the upper-level hierarchy in order to represent a location using relatively small location vocabularies.

\section{Conclusions}
\;\;\;\;\;\;This paper proposed a contextual location embedding model to efficiently handle numerous location vocabularies in various real-world applications. 
We represented a particular location as a combination of several grids at different scales to reduce the number of locations to be trained.
In addition, to incorporate various location functionalities, our model dynamically calculated the embedding vector of a target location, which varies depending on its trajectory.
We employed a variant of the ALM objective, which trains the model with several ALM objectives sequentially. 
The experimental results demonstrated that our model significantly improved the performance of downstream models with fewer model parameters, compared to the existing location embedding methods.

\clearpage
\subsubsection{Acknowledgment}

This work was supported by the institute of Information \& communications Technology Planning \& Evaluation (IITP) grant funded by the Korea government (MSIT) (No.2019-0-00075, Artificial Intelligence Graduate School Program (KAIST)) and the National Research Foundation of Korea (NRF) grant funded by the Korea government (MSIT) (No. NRF-2022R1A2B5B0 \\ 2001913).

\noindent The authors would like to thank the AI Service Business Division of SK Telecom for providing GPU cluster support to conduct massive experiments.

\section*{Ethical Statement}
There are no ethical issues.

\bibliographystyle{splncs04}
\bibliography{mybibliography}

\clearpage
\appendix

\section{Appendix} \label{sec:Appendix}
\subsection{Dataset Details} \label{ssec:Details of Dataset}
\;\;\;\;\;\;We used two real-world datasets: (1) Mobile-T and (2) Geo-Life \cite{zheng2010geolife}.
In this section, we describe the details of pre-processing in two datasets.

\subsubsection{Mobile-T}
This data is a set of user trajectories collected by the base stations of the major cellular network operator. 
Each base station provides a signal to the surrounding area and records the user’s access to the corresponding area.
The average density of base stations in this dataset is about 100m, and therefore we converted the location records in Mobile-T into a grid at a 100m scale.
Mobile signaling datasets are more suitable for evaluating the effectiveness of our model because they contain dense trajectories, unlike some public check-in datasets \cite{an2017poi2vec,lin2020pre}.
Since base stations are densely installed, the signaling data is able to represent the user's overall trajectories.

We randomly sampled about 0.4 milion customers who agreed to collect and analyze their information.
The Mobile-T dataset consists of mobile signaling data; not all location records indicate a user's visit. 
Location records simply passed by the user do not imply explicit purposes.
To filter out such points, we removed the location records which had an average duration time below five minutes.
Then, we calculated the velocity of each location record and denoted \textit{stop} to the location records under 4km/h velocity.
A sequence of all locations between  successive \textit{stop} records is considered to be the user trajectory.
Finally, we derived trajectories from more than ten location records.

Meanwhile, the Mobile-T dataset contains the land usage of the last location of a trajectory.
There are 15 unique land usages of a trajectory, which indicates Apartment House (30.34\%), Factory (2.07\%), Educational Research Facilities (1.53\%), Detached House (36.21\%), Hotel Facilities (0.50\%), Business Facilities (5.38\%), Sports Facilities (0.09\%), Transportation Facilities (0.41\%), Medical Facilities (017\%), Automobile related Facilities (0.49\%), Residential Neighborhood Facilities/class1 (6.90\%), Residential Neighborhood Facilities/class2 (13.90\%), Religion Service Facilities (0.14\%), Storage Facilities (0.49\%), and Shopping Service Facilities (1.06\%).

\subsubsection{Geo-Life}\footnote[3]{https://www.microsoft.com/en-us/download/confirmation.aspx?id=52367}
The trajectories in this dataset are represented as sequences of locations, each of which contains  latitude, longitude, and altitude. 
GeoLife contains 17,621 trajectories collected by 182 users over a period of five years in Microsoft Research Asia.
Among them, trajectories of 73 users have their transportation modes.
The GPS trajectories in this dataset were recorded in every 1-5 seconds, and we selected location records of 1-minute increments.
In addition, we extracted trajectories from more than ten location records.
The way to decompose the location with several hierarchies was the same as that used for the Mobile-T dataset, using latitude and longitude.
The Geo-life dataset contains five unique transportation modes of a trajectory, which indicates bus (18.58\%), car (21.46\%), walk (27.27\%), bike (18.58\%), and subway (7.13\%).

\subsection{Pre-trained Location Embedding model Details} \label{ssec:Details of baselines}
\;\;\;\;\;\;We demonstrated the superiority of our pre-trained location embedding model by comparing six distributed embedding models.

\noindent\textbf{(1) SERM} \cite{yao2017serm}: 
This model is a randomly initialized embedding layer to produce input vectors for downstream task models.
The embedding layer consists of the embedding for the location record, the timestamp, and the text information aligned with the location.
A specific model (e.g., LSTM) for a downstream task is connected to this embedding layer and trained together.
The dimension of the embedding layer was set to 256.
We removed the embedding module for the text information in the original SERM, because there is no a text message that describes the user’s activity in each GPS record of our datasets.

\noindent\textbf{(2) HIER} \cite{shimizu2020learning}: 
The large location vocabulary problem can be solved by reconstructing a location with multiple grids at different scales, and having each large scale grid share the small scale grids.
HIER \cite{shimizu2020learning} decomposes a location into multiple spatial scales to account for the spatial hierarchy in the location embeddings.
In this model, we set the decomposed spatial scales to 100km, 1km, 100m, as in our model.

\noindent\textbf{(3) DeepMove} \cite{zhou2018deepmove}: 
They applied the Skip-gram of Word2Vec to trajectory data which have a set of origin and destination records. We modified the proposed module to fit our data with dense locations in a trajectory. 
The skip-gram with negative sampling was used as a training method, and the window size was set to five. 
The dimension of the location embedding in DeepMove was set to 256.

\noindent\textbf{(4) TALE} \cite{wan2019learning}: 
The CBOW module of Word2Vec was employed to generate pre-trained vectors of  locations. To reduce computational complexity, they used the hierarchical softmax method for training, but we instead employed negative sampling to improve performance. The rest of the parameter settings were identical to the DeepMove.

\noindent\textbf{(5) CTLE} \cite{lin2020pre}: They used the bidirectional transformer encoder architecture with Masked Language Model (MLM) pre-training objective to derive the context-aware location embedding vectors, considering contexts.
In this model, six stacks of Transformer encoder layers which contained eight attention heads are employed, and the dimension of the embedding layer and final location embedding vector was set to 256.

\noindent\textbf{(6) TrajFormer} \cite{liang2022trajformer}: They developed the squeezed Transformer Encoder to classify the transportation modes of a trajectory, effectively diminishing the dimensions of keys and values prior to computing the self-attention module.
In this model, six stacks of Transformer encoder layers which contained eight attention heads are employed, and the dimension of the embedding layer and final location embedding vector was set to 256.
The squeeze rate is set to $1$ for the best performance.
In our experiments, the sub-path labeling in this model was removed.

\begin{table}[]
\centering
\caption{Hyperparameters of our location pre-trained embedding model on Mobile-T and Geo-Life.}
\label{tab:Hyperparameters}
\begin{tabular}{c|cc}
\hline
\hline
\textbf{Hyperparameter}                                                  & \textbf{Mobile-T} & \textbf{Geo-Life} \\ \hline
Epoch                                                                    & 20                & 10                \\
Batch size                                                               & 32                & 32                \\
Hidden size                                                              & 256               & 256               \\
Attention dropout                                                        & 0.1               & 0.1               \\
\# heads                                                                & 8                 & 8                 \\
\begin{tabular}[c]{@{}c@{}} \# transformer\\ decoder layers\end{tabular} & 6                 & 6                 \\
Max sequence length T                                                    & 32                & 32                \\
Hierarchy Level                                                          & 3                 & 3                 \\
Adam $\epsilon$                                                                     & 1e-4              & 1e-4              \\
Adam ($\beta_{1},\beta_{2}$)                                                                    & (0.9, 0.999)      & (0.9, 0.999)      \\
Weight decay                                                             & 1e-2              & 1e-2              \\
\# warm-up steps                                                        & 10000             & 10000             \\ \hline \hline
\end{tabular}
\end{table}

\subsection{Pre-training Details} \label{ssec:Training Details}
\;\;\;\;\;\;Table \ref{tab:Hyperparameters} describes the optimal hyperparameters of our location pre-trained embedding model. 
The max length of an input trajectory was set to 32, and the batch size was 32.
Each dimension of the embedding layers in the Geo-tokenizer embedding layer ($z$) was set to 256, and the dimensions of the final embedding vectors in the pre-trained location embedding models ($u$) was set to 256.
Our model adopted six stacks of transformer dncoder layers which contained eight attention heads.
The number of layers in the feed-forward network of HALM is two. 
The Adam optimizer with a learning rate of 0.001, $\beta_1$ of 0.9, and $\beta_2$ of 0.999 was used to find the optimal parameters of our model.
We trained our pre-training and fine-tuning models with Cross-entropy loss.
Our model was trained using one V100 GPU.

\subsection{Fine-tuning Downstream model Details} \label{ssec:Details of Fine-tuning}
\subsubsection{Next Location Prediction task}
Given a trajectory $s'=\{l_0,l_1,...,l_T\}=\{(l^1_0, l^2_0, ..., l^H_0)$
$,(l^1_1, l^2_1, ..., l^H_1),\ldots,(l^1_T, l^2_T, ..., l^H_T)\}$, the downstream model connected to our pre-trained location model $u$ is a function $g$ to predict the next location $l_{T+1}=(l^1_{T+1}, l^2_{T+1}, ..., l^H_{T+1})$ as shown in Figure \ref{fig:finetune_model}a.
Similar to the pre-training stage, the fine-tuning for the next location prediction is a multi-task model, which predicts the $H$ next location components of all hierarchies, respectively.
We consider the next prediction to be correct when all the hierarchies $(l^1_{T+1},l^2_{T+1},...,l^H_{T+1})$ are simultaneously correct.
Therefore, the function $g$ contains $H$ independent layers to predict the $H$ next location components of $H$ hierarchies.
Then, the probability of the next location component in the level-$h$ hierarchy is calculated in the same way as the HALM objective in pre-training.
In this paper, we employed two models as the function $g$: (1) a fully-connected layer and (2) LSTM (Long-Short Term Memory) \cite{hochreiter1997long}.

\textbf{(1) FFN}: 
The prediction of the next location $l^h_{T+1}$ is sequentially implemented using a fully-connected feed-forward network $g$ as follows:
 \begin{equation}
 \begin{split}
 \label{equation:ffn_fully}
    & \widehat{l^h_{T+1}}=g_{h}(\frac{1}{T}\sum_{t=1}^{T}(\mathbf{e}_t^{(N)}) \mathbin\Vert \widehat{o^{h-1}_{T+1}}), \\
    & \widehat{o^{0}_{T+1}}=\mathbf{0},
 \end{split}
 \end{equation}
where $\mathbin\Vert$ is the concatenation operation, $\widehat{o^{h-1}_{T+1}}$ is the one-hot encoding vector from the prediction result $\widehat{l^{h-1}_{T+1}}$, and $g_h$ is the feed-forward network of the level-$h$ hierarchy, used to predict the next location component $l^h_{T+1}$. 
The $\mathbf{e}_t^{(N)}$ is the pre-trained model's output  vector corresponding $t$-th step.
In short, this model is a fully-connected feed-forward network, which uses the average of the output vectors of pre-trained location embeddings model and predicts the location of the $T+1$ timestamp.
For the experiments, the model $g$ consists of one fully-connected feed-forward layer, and the output of the model $g$ are fed into a softmax layer.

\textbf{(2) LSTM}:
The prediction of the next location $l^h_{T+1}$ is sequentially implemented using LSTM layer $g$ as follows:
 \begin{equation}
 \begin{split}
 \label{equation:ffn_h_lstm}
    & \widehat{l^h_{T+1}}=g_{h}([\mathbf{e}_{t}^{(N)} \mathbin\Vert \widehat{o^{h-1}_{T+1}}]_{t\in\{1:T\}}), \\
    & \widehat{o^{0}_{T+1}}=\mathbf{0},
 \end{split}
 \end{equation}
where $\widehat{o^{h-1}_{T+1}}$ is the one-hot encoding vector from the prediction result $\widehat{l^{h-1}_{T+1}}$, $g_h$ is the LSTM layer of the level-$h$ hierarchy, and $\mathbf{e}_{t}^{(N)}$ is the pre-trained model's output  vectors in the $t$-th step.
We use the output representation of the last output state of the LSTM to predict the next location component $l^h_{T+1}$. 
In short, the sequence of output vectors of the pre-trained location embedding model were sequentially fed into the one LSTM layer and the softmax layer, to predict the location of the $T+1$ timestamp considering temporal correlation.

\subsubsection{Classification task}
In this task, given a trajectory $s'=\{l_0,l_1,...,l_T\}=\{(l^1_0, l^2_0, ..., l^H_0),(l^1_1, l^2_1, ..., l^H_1),$
$\ldots,(l^1_T, l^2_T, ..., l^H_T)\}$, the downstream model connected to our pre-trained model $u$ is a function $q$ to classify the land usages or transportation modes as shown in Figure \ref{fig:finetune_model}b.
The average of the output vectors of the pre-trained location embeddings model of time $t$s is fed into the function $q$.
The model $q$ consists of one fully-connected feed-forward layer and the softmax layer.

\subsection{Extended Study on the components (RQ2)} \label{subsection:extended_ablation_study}
\begin{figure*}[h]
    \centering
    \includegraphics[width=0.9\linewidth]{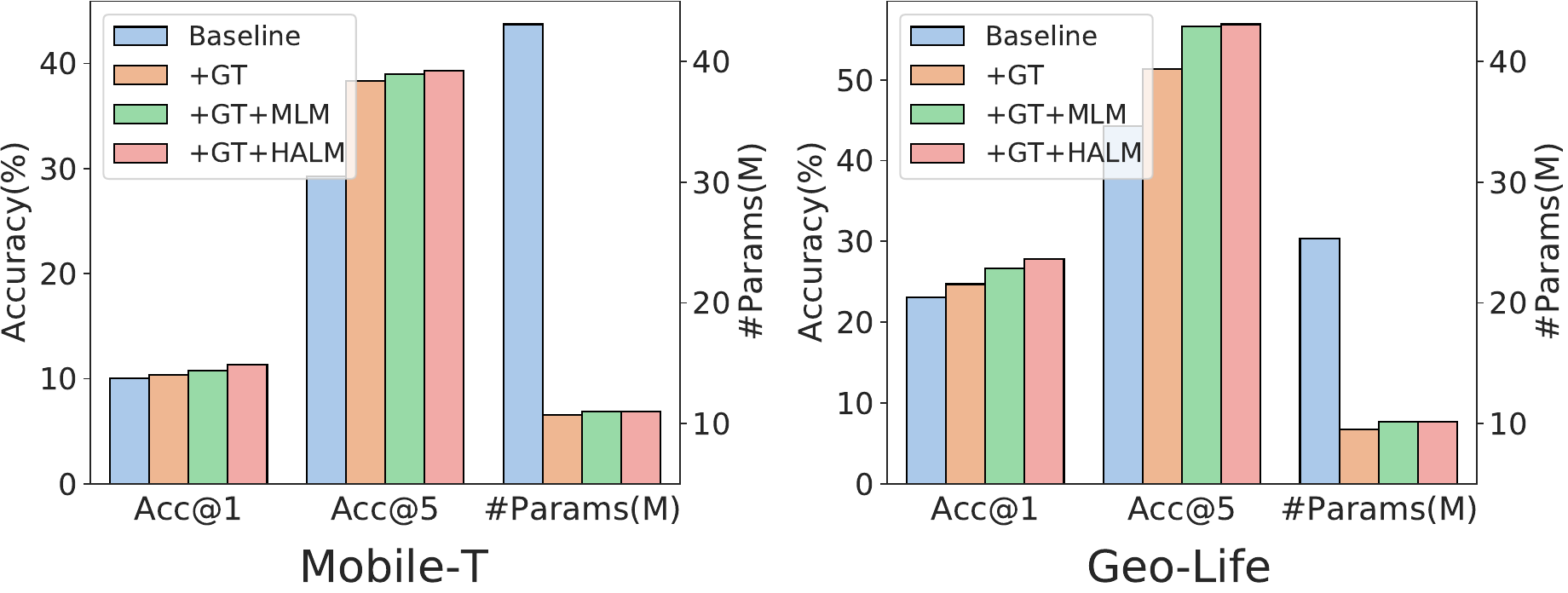}
    \caption{Comparison of next location prediction performance and efficiency for different combinations of components. 
        }
    \label{fig:ablation_2_1}      
\end{figure*}

 \;\;\;\;\;\;As shown in Figure \ref{fig:ablation_2_1}, we further investigated the effectiveness of each component of our pre-trained location embedding model by designing four variants as follows:

\noindent\textbf{(1) Baseline}: 
This model utilizes the original transformer decoder using the ALM objective for pre-training without the Geo-tokenizer embedding layer.
This is a simple auto-regressive pre-trained model.

\noindent\textbf{(2) +Geo-tokenizer(GT)}: 
This model replaces the embedding layers in the baseline with the Geo-tokenizer embedding layer, which decomposes each location record into the three hierarchical components (100km, 1km, 100m). 
The pre-trained model's objective is the basic ALM proposed in the transformer \cite{vaswani2017attention}.
Therefore, the ALM objectives of the three hierarchies are independent.

\noindent\textbf{(3) +Geo-tokenizer(GT)+MLM}: 
This model uses the Geo-tokenizer fused on the baseline and employs the Masked Location Model (MLM) objective of the CTLE\cite{lin2020pre}.
This is the pretrained model replacing our causal location embedding model with the stack of the bidirectional transformer encoders.

\noindent\textbf{(4) +Geo-tokenizer(GT)+HALM}: 
This model uses the Geo-tokenizer fused on the baseline and employs the HALM objective.
This is our proposed model.
This shows that our proposed HALM method is superior to the MLM method.

\vspace{-0.5cm}
\subsection{Our Deployed Solution} \label{sec:Our Deployed Solution}
\begin{figure}[ht]
    \begin{center}
    \includegraphics[width=0.6\linewidth]{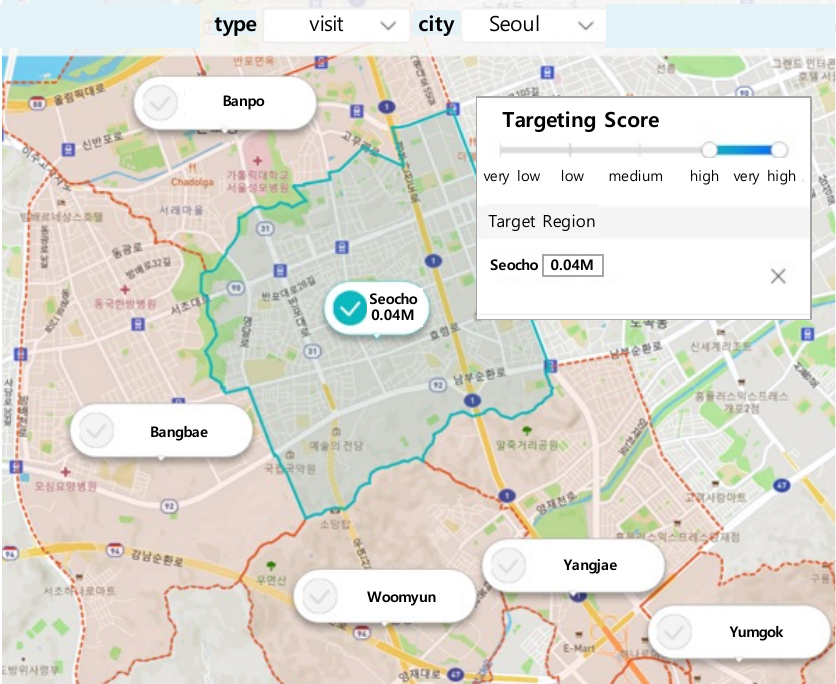}
    \end{center}
    \caption{Illustration of our location-based marketing tool. Seocho refers to a district in Seoul, Korea. In this figure, the estimated number of people expected to visit the Seocho district is 0.04 million. Using this tool, we can identify those who are likely to visit the Seocho district. The abbreviation \textbf{M} represents million.}
    \label{fig:deployed_solution}
    \end{figure}

\;\;\;\;\;\;Our pre-trained model has been implemented in an inter-company marketing tool, designed to predict individuals likely to visit a particular area for location-based marketing purposes. The deployed solution effectively encompasses entire regions within the author's country by utilizing a next-location prediction model built upon our pre-trained model.
Figure \ref{fig:deployed_solution} shows a screenshot of our custom GUI tool, which extract the list of customers who will move to a specific region given his/her trajectory.

\subsection{Qualitative analysis}
\begin{figure}
    \begin{center}
    \includegraphics[width=0.9\linewidth]{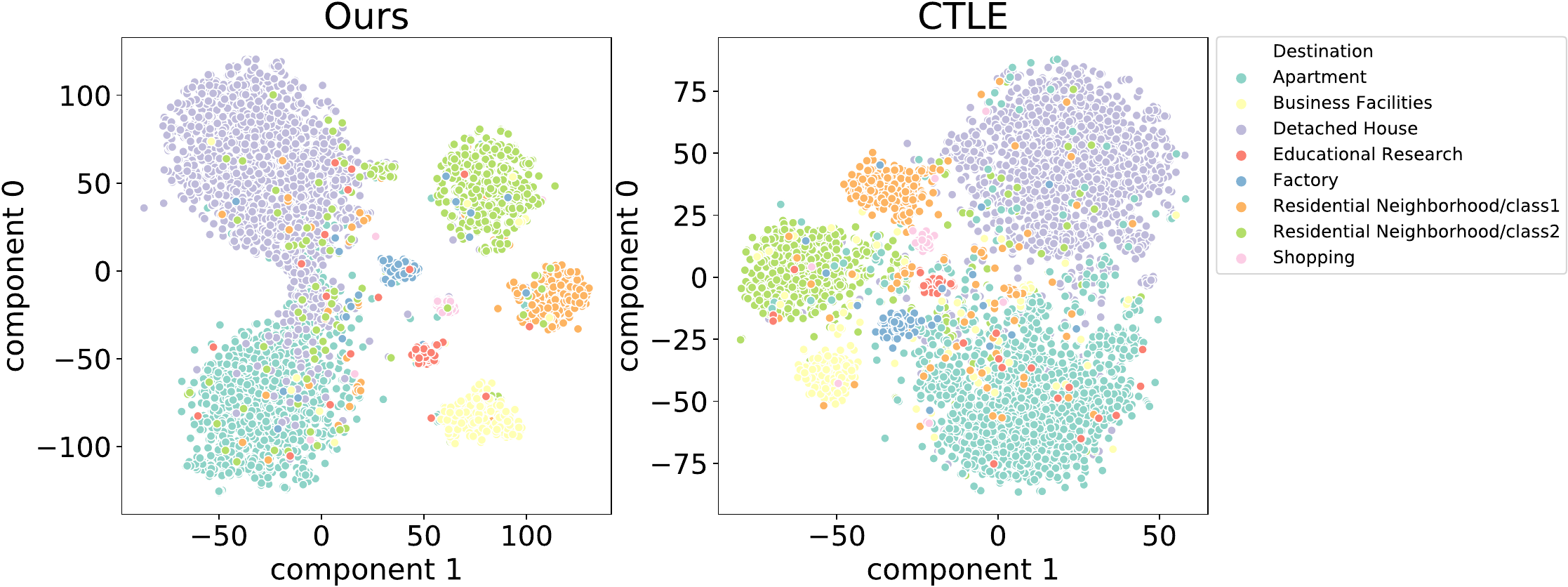}
    \end{center}
    \caption{Visualization of the pre-trained trajectories' representations by our model and CTLE on the Mobile-T dataset.}
    \label{fig:clustering}
    \end{figure}

\;\;\;\;\;\;We also compared our pre-trained model with CTLE by visualizing the trained trajectories' representations ($\sum_{t=1}^{T}e_{t}^{N}$) using t-SNE \cite{van2008visualizing} from the test dataset (Mobile-T), as shown in Figure \ref{fig:clustering}.
CTLE is the state-of-the-art model for several downstream tasks such as the next location prediction.
In the Mobile-T dataset, the land usage of the last location is the purpose of the trajectory (i.e., destination).
It can be seen that trajectories' representations trained by our model tend to push trajectories of different purposes than CTLE.
It reflects that representations learned by our model can capture the semantic purpose of the trajectory.

\end{document}